\documentclass{article}

\usepackage[preprint]{neurips_2026}
\usepackage{amsmath}
\usepackage{amssymb}
\usepackage{mathtools}
\usepackage{amsthm}

\usepackage[utf8]{inputenc} %
\usepackage[T1]{fontenc}    %
\usepackage[hidelinks]{hyperref}       %
\usepackage{url}            %
\usepackage{booktabs}       %
\usepackage{amsfonts}       %
\usepackage{nicefrac}       %
\usepackage{microtype}      %
\usepackage{xcolor}         %
\usepackage{wrapfig}
\usepackage{graphicx}
\title{X-AVDD: Cross-Attentive Audio-Visual Dataset Distillation}

\author{
Saumyaranjan Mohanty$^{1}$ \quad
Aravind Reddy$^{2}$ \quad
Konda Reddy Mopuri$^{1}$\\
{\small $^{1}$Department of Artificial Intelligence, Indian Institute of Technology Hyderabad}\\
\resizebox{\dimexpr\textwidth-2\tabcolsep\relax}{!}{\small $^{2}$Centre for Responsible AI, Wadhwani School of Data Science \& AI, Indian Institute of Technology Madras}\\
{\tt\small ai23resch04001@iith.ac.in \quad aravind@cerai.in \quad krmopuri@ai.iith.ac.in}
}

\hypersetup{pdftitle={X-AVDD: Cross-Attentive Audio-Visual Dataset Distillation},pdfauthor={Saumyaranjan Mohanty, Aravind Reddy, Konda Reddy Mopuri}}

\begin{document}

\maketitle

\begin{abstract}
\textit{Audio-visual dataset distillation} aims to synthesize a small dataset from a large paired audio-visual dataset, while nearly preserving the training utility for neural networks. Existing approaches such as DM and AVDD face two critical limitations: 1) they are memory-intensive, making them difficult to scale to complex architectures and higher samples-per-class (SPC) settings, and 2) they often fail to preserve strong cross-modal semantic correspondence. To tackle these issues, we propose \textbf{X-AVDD}, a \textbf{cross-attention-based} distillation method that explicitly captures inter-modal relationships during synthesis. By decoupling cross-modal alignment from joint optimization, X-AVDD is substantially more memory-efficient, reducing end-to-end synthesis cost by $\sim 8\times$ GPU-hours, while producing highly aligned synthetic data. Empirically, X-AVDD significantly outperforms other audio-visual distillation methods. On VGGSound, an audio-visual dataset of short YouTube clips, X-AVDD improves top-1 test accuracy from $8.2\%\rightarrow24.0\%$ at SPC=$10$ and $9.8\%\rightarrow26.5\%$ at SPC=$20$. Furthermore, X-AVDD improves cross-architecture generalization (e.g., $25.5\%\rightarrow35.3\%$ for a ViT target on AVE dataset at SPC $=50$) and successfully scales to complex architectures such as AudioCLIP, achieving $62.3\%$ accuracy using only $5.65\%$ of the full dataset, against the full dataset accuracy of $72.4$\% while completing in $9\times$ lower training time compared to the full dataset.
\end{abstract}
    
\section{Introduction}
\label{sec:intro}
\begin{wrapfigure}{r}{0.46\textwidth}
  \centering
  \includegraphics[width=\linewidth]{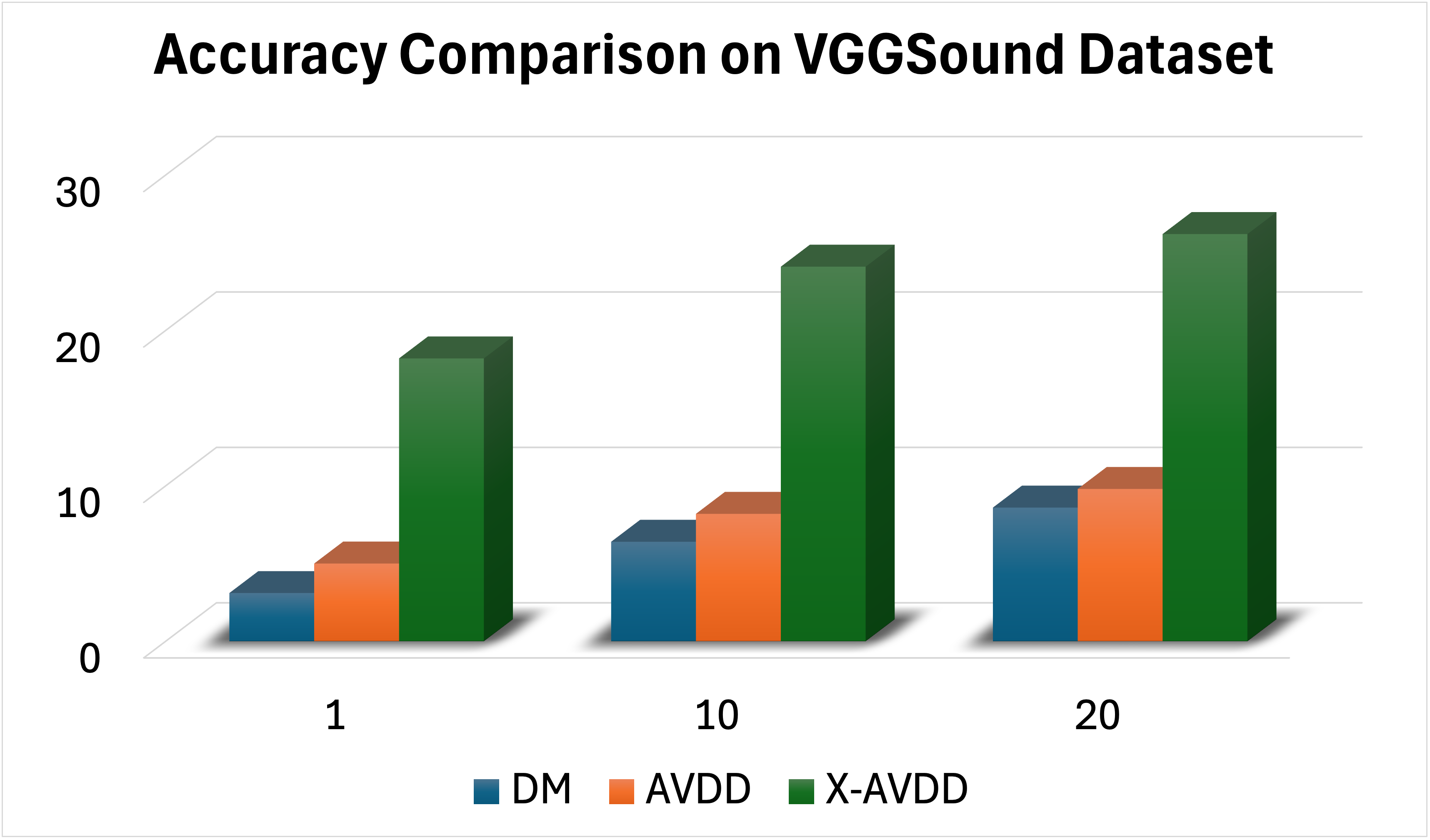}
  \caption{Accuracy on the VGGSound dataset under different samples-per-class settings. X-AVDD consistently outperforms DM and AVDD~\citep{kushwaha2024audiovisual}.}
  \label{fig:aaai_vggsound_compare}
\end{wrapfigure}

Modern multimodal models derive much of their capability from training on massive datasets that jointly capture complementary signals across modalities (e.g., vision, audio, text)~\citep{11027559}.
While such datasets enable richer representations and improved robustness over single-modality learning, their scale creates a growing bottleneck: storing, curating, and repeatedly training on multimodal corpora is expensive in both computation and energy~\citep{schwartz2020greenai,strubell2019energy}.
Recent work has therefore increasingly emphasized \emph{data-centric} methods to improve data quality, and data curation as a key lever for efficiently scaling multimodal models~\citep{11027559}.

\textit{Dataset distillation} (DD), also known as Dataset Condensation, is an increasingly popular route to data-efficient learning. DD methods synthesize a small set of labeled synthetic examples such that models trained on the synthetic data achieve performance comparable to those trained on the full dataset~\citep{wang2018dataset,zhao2021dataset,sachdeva2023data, liu2025evolutiondatasetdistillationscalable}. Distilled datasets enable rapid iteration for tasks like neural architecture search and hyperparameter tuning~\citep{ding2024calibrateddatasetcondensationfaster,such2020generative}, and provide a compact interface for sharing training signals while maintaining privacy~\citep{dong2022privacy}.

As DD moves beyond unimodal settings (primarily image classification), \emph{multimodal} distillation~\citep{wu2024visionlanguage} introduces a central additional challenge: the synthetic data must preserve not only modality-specific information, but also the cross-modal associations that make multimodal learning effective. In this paper, we focus on the multimodal distillation problem of \textit{audio-visual dataset distillation} introduced in~\citep{kushwaha2024audiovisual}. Given a large-scale paired audio-visual dataset, we want to synthesize a much smaller dataset such that downstream models trained on the synthetic dataset have good performance. Following~\citep{kushwaha2024audiovisual}, we consider audio-visual event recognition, which is the closest audio-visual analogue of image classification, as the main proxy task for model evaluation. Their method AVDD, extends Distribution Matching (DM)~\citep{10030751} to paired audio-visual data. However, their approach faces two fundamental limitations: 

\noindent \textbf{Memory Inefficiency.} It is highly memory-intensive, making it difficult to scale distillation to higher samples-per-class (SPC) settings and complex modern architectures. 

\begin{figure}[hbt!]
  \centering
  \includegraphics[width=0.6\linewidth]{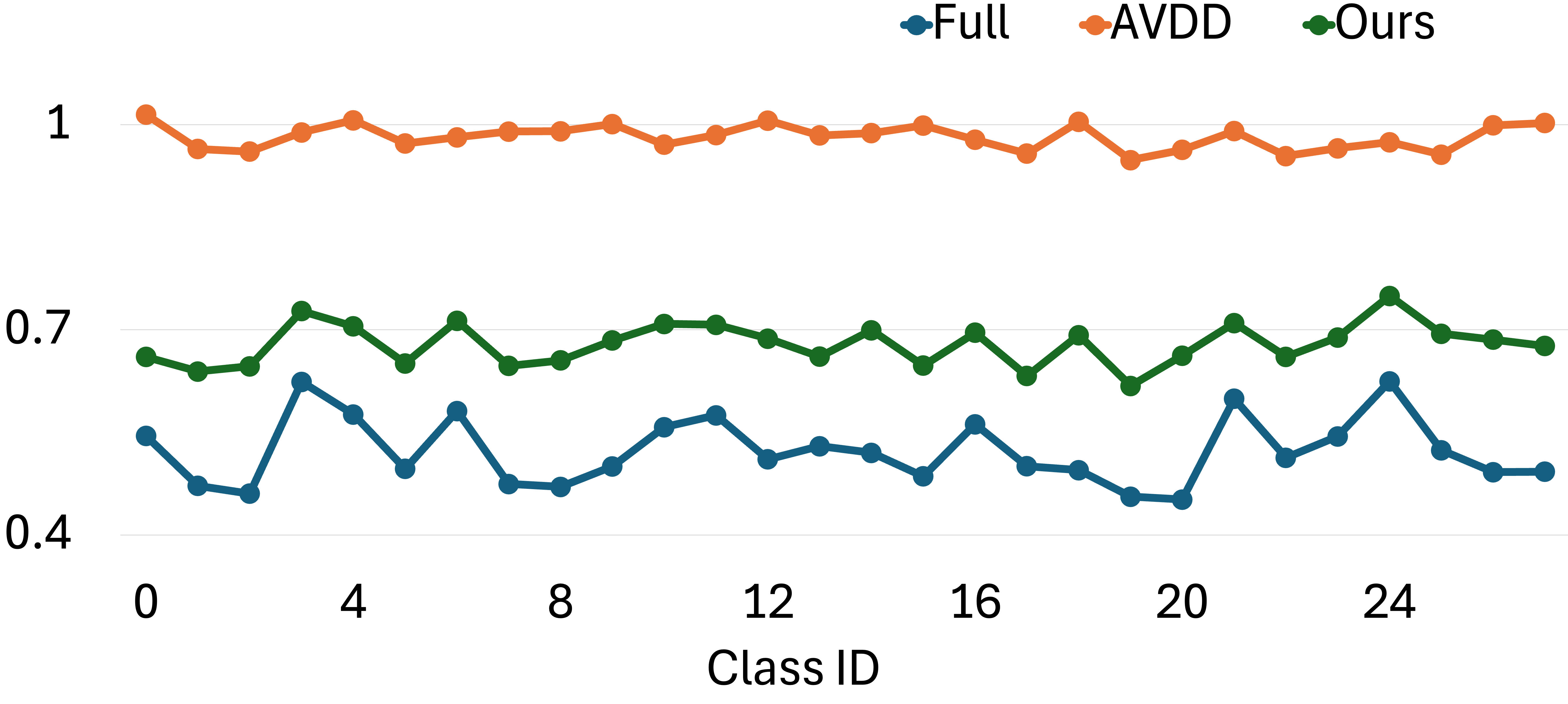}
  \caption{Modality gap (class-wise $\ell_2$ distance between audio and visual class centers in the feature space) on the AVE dataset, and datasets distilled by AVDD and X-AVDD using AudioCLIP network. X-AVDD achieves a consistently lower modality gap.}
  \label{fig:modality_gap}
\end{figure}

\noindent \textbf{Modality Gap.} It fails to preserve strong semantic correspondence between modalities, leading to a high ``modality gap''~\citep{liang2022mind,zhang2025beyond}. As shown in Figure~\ref{fig:modality_gap}, AVDD generates synthetic data with significantly higher class-wise $l_{2}$ distances between audio and visual features compared to the original dataset.

To address these issues, we propose \textit{X-AVDD}, a \emph{cross-attentive} audio-visual dataset distillation method designed for scalability and stronger cross-modal coupling. The key idea is to \emph{explicitly capture inter-modality relationships during distillation} using bi-directional, multi-head cross-attention, while \emph{decoupling cross-modal alignment from joint optimization} via a pre-trained teacher model. 

This design substantially improves efficiency: X-AVDD reduces total synthesis time on a Tesla V100 GPU from \textbf{10 hours} to \textbf{1.3 hours}. Figure~\ref{fig:aaai_vggsound_compare} presents the impressive generalization performance obtained by X-AVDD compared to the baseline DM and AVDD methods.

\paragraph{Contributions.}
\begin{itemize}
    \item We introduce \textbf{X-AVDD}, a cross-attention-based method that explicitly models inter-modal relationships during audio-visual dataset distillation.
    \item We propose a \textbf{decoupled cross-modal alignment strategy} using bi-directional cross-attention, yielding substantially improved \textbf{memory efficiency} and enabling scaling to larger SPC and modern architectures.
    \item We demonstrate that X-AVDD produces \textbf{better-aligned synthetic data} (lower modality gap) and delivers consistent improvements over AVDD and multimodal SRe$^2$L baselines across multiple benchmarks.
\end{itemize}

\section{Related Works}
\subsection{Dataset Distillation}
Dataset Distillation (DD), also referred to as Dataset Condensation, aims to synthesize a compact set of labeled examples that preserves the training signal of a much larger original dataset. While early methods rely on bi-level optimization~\citep{wang2018dataset} (e.g., gradient matching~\citep{zhao2021dataset,zhao2021siamese}, trajectory matching~\citep{cazenavette2022distillation,cui2023scaling}), which are quite memory and compute intensive. To improve scalability, distribution-matching methods~\citep{10030751} align the statistics of real and synthetic embeddings. However, these methods often encounter memory bottlenecks as the synthetic budget or model complexity increases. We refer the reader to recent surveys for broader coverage and taxonomies of DD methods~\cite{sachdeva2023data,lei2023comprehensive,liu2025evolutiondatasetdistillationscalable}.

\subsection{Attention in DD}
While attention has been explored in unimodal distillation, such as DataDAM~\citep{sajedi2023datadam}, which matches spatial attention maps, our work is the first to employ bi-directional cross-attention in a multimodal context. Unlike DataDAM's focus on unimodal spatial patterns, X-AVDD uses bi-directional multi-head cross-attention to explicitly model the temporal and semantic dependencies between audio and visual streams during synthesis. 

\subsection{Multimodal Dataset Distillation (MDD)}
MDD extends distillation to settings requiring the preservation of cross-modal associations. While vision-language distillation~\citep{wu2024visionlanguage,xu2024lowrank,zhang2025beyond,zhao2025efficient,lee2025covmatch} has seen rapid progress, audio-visual distillation remains under-explored. A primary challenge in MDD is the ``modality collapse'', in which synthetic pairs lose semantic alignment~\citep{zhang2025beyond}. Our work is the first to address this in the audio-visual domain by using a decoupled alignment strategy to proactively minimize this gap.  

\subsection{Audio-Visual Event Recognition}
Recognizing events in unconstrained videos requires learning aligned representations across audio and visual modalities~\citep{BROUSMICHE202252}. Benchmark datasets such as AVE~\citep{Tian_2018_ECCV} and VGGSound~\citep{9053174} have driven the development of complex architectures, such as AudioCLIP~\citep{9747631}. Contrastive learning and cross-modal attention have become prominent directions for learning aligned representations across modalities~\citep{8237335,miech2019howto100mlearningtextvideoembedding}. However, the computational cost of training these models on the full dataset is quite high. Our work contributes to this domain by making audio-visual training more data-efficient through distillation, enabling faster iteration and cheaper training in this setting. 

\subsection{Limitations of AVDD}
Despite establishing the first framework for audio-visual dataset distillation, AVDD faces a fundamental scalability bottleneck due to its reliance on a non-decoupled Distribution Matching (DM) paradigm. In AVDD, the entire set of synthetic samples must be held in GPU VRAM simultaneously during the optimization loop to match the global feature distributions, leading to Out-of-Memory (OOM) failures as the SPC budget or model complexity increases.  While a class-wise batching strategy could theoretically reduce memory, our empirical tests show that it drastically degrades performance (e.g., a drop from $58.0\%$ to $49.5\%$ at SPC=$20$), because it loses the inter-class relationship information necessary for effective distribution matching. Consequently, AVDD is largely restricted to simple ConvNet architectures and lower SPC settings, failing to generalize to modern high-parameter models like ViT~\citep{dosovitskiy2021an} or AudioCLIP~\citep{9747631}. Furthermore, AVDD involves a ``buffer creation'' cost (reflected in our AVDD timing breakdown in Table~\ref{Tab:memory_compare}).

X-AVDD addresses these limitations by decoupling the alignment phase from the synthesis loop, enabling class-wise optimization via frozen, cross-attentive-guided networks that maintain superior accuracy while reducing peak VRAM usage and total synthesis time.

\section{Methodology}
\label{sec:methodology}
In this section, we describe our proposed X-AVDD in detail. Table~\ref{table:notations} (in Appendix A) provides the notation used throughout the paper. 

\textbf{Problem formulation}.
Let the labeled dataset be $\mathcal{V} = \{(x_i^a, x_i^v, y_i)\}_{i=1}^{N}$, where $x_i^a \in \mathcal{X}^a$ and $x_i^v \in \mathcal{X}^v$ denote the audio and visual modalities of the $i$-th sample, respectively, and $y_i \in \mathcal{Y}$ is the corresponding ground-truth label. The visual input $x_i^v$ is represented by the central frame of the video, while the audio input $x_i^a$ is represented by the corresponding spectrogram.

The goal of DD is to learn a compact synthetic dataset $\mathcal{S}=\{(\tilde{x}_i^a, \tilde{x}_i^v, \tilde{y}_i)\}_{i=1}^{M}, M \ll N$, such that a model parameterized by $\theta^{\mathcal{S}}$, trained on $\mathcal{S}$, achieves generalization performance comparable to a model $\theta^{\mathcal{V}}$ trained on the full dataset $\mathcal{V}$.

\textbf{Cross Attention}. This is a fundamental attention mechanism, which allows different modalities to exchange information in a multi-modal setting~\citep{tsai-etal-2019-multimodal}. Information exchange occurs through queries from one modality attending to keys and values from another modality. Cross-attention enables fine-grained semantic alignment across different input streams. Intuitively, for audio-visual dataset distillation, cross-attention can capture complementary information~\citep{9856650,10.1145/3606843.3606852} between the two modalities and guide distillation steps towards generating synthetic samples that capture informative semantic information from the real dataset.

\begin{figure}[hbt!]
    \centering
    \includegraphics[width=0.8\textwidth, trim={0 0 0 0}, clip]{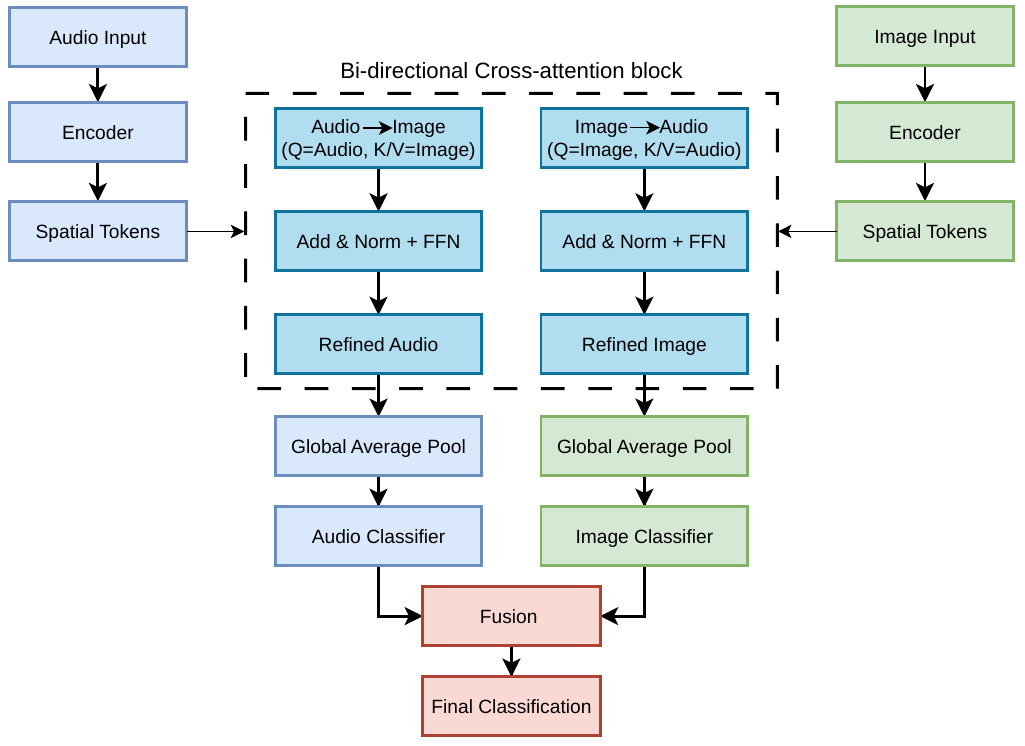}
    \caption{Overview of the bi-directional cross-attention mechanism for Audio-Visual event recognition.}
    \label{fig:bi_directional}
\end{figure}

Some recent works~\citep{hiller2024bixt,si2025bidirectional} have explored bidirectional cross-attention for training models on multimodal datasets. Bi-directional cross-attention is a mechanism in multimodal models in which two modalities (e.g., text and images, or audio and video) attend to each other in both directions, allowing each to refine its representation based on the other. Concretely, one attention pass lets modality A query and extract relevant information from modality B, while a second, symmetric pass lets modality B query modality A. This mutual exchange captures richer, more aligned relationships than one-way fusion. This bidirectional interaction improves tasks such as alignment, retrieval, and generation by modeling inter-dependencies rather than treating one modality as dominant, leading to more coherent, semantically grounded multimodal representations. Figure~\ref{fig:bi_directional} shows the block diagram of the implementation of the bi-directional cross-attention mechanism. 

\textbf{Formulation of X-AVDD}. In this paper, we propose \textit{X-AVDD}, a cross-attention-based decoupled dataset distillation method. Our proposed method decouples distillation of synthetic data from training of teacher model with original dataset. This decoupling ensures distillation process can be parallelized and does not depend upon having access to the original large dataset. 
We incorporate a bidirectional cross-attention module that enables the extraction of complementary information from both modalities. Our method consists of three steps. 
\begin{enumerate}
    \item Pre-training of the teacher network, which incorporates bi-directional cross-attention.
    \item Computation (and storage) of real embeddings for both audio and visual modalities through the cross-attention layers.
    \item Iterative learning of a synthetic dataset by matching cross-attended embeddings from real and synthetic samples.
\end{enumerate}

\vspace{-2mm}

In our approach, we use randomly sampled audio and frame pairs from real data to initialize the distillation process (similar to~\citet{cazenavette2022distillation} and ~\citet{du2024diversity}). For the experiments, the same set of initialization samples is used for both the baseline and our proposed method. We use a Euclidean Distribution Matching (EDM) loss function that encourages synthetic embeddings to cluster near real ones in Euclidean space by minimizing the pairwise Euclidean distances between attended embeddings from real and synthetic samples. Pairwise Euclidean distance between two matrices $\mathbf{A}$ and $\mathbf{B}$ of dimensions $(M,d)$ and $(N,d)$ respectively, is calculated as: 
\begin{equation}
\displaystyle D_{\text{euc}}(\mathbf{A},\mathbf{B}) = \sum_{j=1}^{N}{\sum_{i=1}^{M}\|\mathbf{A}[i,:]-\mathbf{B}[j,:]\|_2}
\label{eq:pairwise_euclidean_distance}
\end{equation}

Let $\{x_{ar}^C,x_{vr}^C\}$ and $\{x_{as}^C,x_{vs}^C\}$ be the real and synthetic audio-image sample pairs belonging to class $C$, respectively. Let $h_a$ and $h_v$ be audio and visual encoders, and $\tau_a$ and $\tau_v$ be the tokenization operators for audio and visual modality, respectively. The spatial tokens for real and synthetic samples are expressed as:
\begin{align}
    \mathbf{A_r^C}=\tau_a(h_a(x_{ar}^C)) & \quad \quad \mathbf{V_r^C}=\tau_v(h_v(x_{vr}^C)) \\
    \mathbf{A_s^C}=\tau_a(h_a(x_{as}^C)) & \quad \quad \mathbf{V_s^C}=\tau_v(h_v(x_{vs}^C))
\end{align}

Let $\mathcal{T}_a$ and $\mathcal{T}_v$ be audio $\rightarrow$ image and image $\rightarrow$ audio branches of the bi-directional cross-attention block, respectively. The output of the Global Average Pool (GAP) is the attended embeddings of corresponding real data, which are stored and utilized during the distillation process. Let $g_a$ and $g_v$ be the GAP operators for the audio and visual branches, respectively. The attended embeddings for real data can be represented as : 
\begin{align}
\mathbf{E}_{r({A\rightarrow V})}^C = g_a(\mathcal{T}_a(\mathbf{A_r^C},\mathbf{V_r^C})) & \quad \quad 
\mathbf{E}_{r({V\rightarrow A})}^C = g_a(\mathcal{T}_v(\mathbf{A_r^C},\mathbf{V_r^C}))
\end{align}

The attended embeddings for the synthetic data are given by:
\begin{align}
\mathbf{E}_{s({V\rightarrow A})}^C = g_a(\mathcal{T}_a(\mathbf{A_s^C},\mathbf{V_r^C})) & \quad \quad 
\mathbf{E}_{s({A\rightarrow V})}^C = g_a(\mathcal{T}_v(\mathbf{A_r^C},\mathbf{V_s^C}))
\end{align}

Note that the audio-guided image embeddings $\mathbf{E}_{s({A\rightarrow V})}^C$ are computed by using synthetic image as query and real audio as key and value. Similarly, the image-guided audio embeddings $\mathbf{E}_{s({V\rightarrow A})}^C$ are computed by using synthetic audio as query and real image as key and value. As stated earlier, in order to encourage synthetic embeddings to
cluster near real ones, we perform Euclidean distribution matching (EDM), defined as: 
\begin{align}
l_{A\rightarrow V}^C = D_{\text{euc}}(\mathbf{E}_{r(A\rightarrow V)}^C,\mathbf{E}_{s(A\rightarrow V)}^C) \\
l_{V\rightarrow A}^C = D_{\text{euc}}(\mathbf{E}_{r(V\rightarrow A)}^C,\mathbf{E}_{s(V\rightarrow A)}^C)
\end{align}
The complete loss function during the synthesis stage is:
\begin{align}
l^C = \lambda_{ce}l_{ce}^{C}+ \lambda_{av}l_{(A\rightarrow V)}^C + \lambda_{va}l_{(V\rightarrow A)}^C
\label{eq:total_loss_syn}
\end{align}

Where $l_{ce}^{c}$ is the standard cross-entropy loss of the classification task, and $\lambda_{ce},\lambda_{av}$, and $\lambda_{va}$ are corresponding weights of the loss functions.

\section{Experiments and Results}
\label{sec:Experiments}
\paragraph{Datasets and protocol.}
We evaluate our method on four benchmark audio-visual datasets: AVE~\citep{Tian_2018_ECCV}, MUSIC-21~\citep{Zhao_2019_ICCV}, VGGSound~\citep{9053174}, and VGGSound-10K (10-class subset released with AVDD~\citep{kushwaha2024audiovisual}). 
We use AVDD’s preprocessing and splits for fair comparison, and report mean and standard deviation of test accuracy over five seeds (0--4). 
See Appendix~\ref{appendix:dataset_details} for dataset and preprocessing details.

\textbf{Backbone architecture}. Following AVDD, we first employ a ConvNet architecture~\citep{zhao2021dataset} for both audio and video encoders. Additionally, we employed more complex architectures, such as ResNet-18~\citep{7780459}, ViT~\citep{dosovitskiy2021an}, and AudioCLIP~\citep{9747631}, to thoroughly evaluate X-AVDD's applicability. 

\textbf{Cross-architecture comparison} To validate the effectiveness of the synthetic dataset, we train randomly initialized models, which are different from the backbone architecture utilized to generate the synthetic images. We used ResNet-18 to generate synthetic images, and VGG-11~\citep{simonyan2015deepconvolutionalnetworkslargescale} and ViT~\citep{dosovitskiy2021an} models for cross-architecture comparison. 

\textbf{Baselines}. For the ConvNet architecture, we compared AVDD-reported results with those obtained by our proposed method. For the other architectures, AVDD results in an OOM issue due to hardware limitations. We instead designed and implemented a new multimodal adaptation of SRe$^2$L~\citep{yinsqueeze}, as a baseline for distillation, and compared it against X-AVDD. Details of the multimodal adaptation of SRe$^2$L are provided in Appendix~\ref{sre2l_implementation}. Hyperparameter details are provided in Appendix~\ref{hyperparameter_details}.

\subsection{Results for ConvNet, ResNet-18 and ViT architectures}
Table~\ref{tab:combined_results} provides test set accuracies obtained by training ConvNet, ResNet-18, and ViT architectures from scratch using the distilled dataset for various Samples Per Class (SPC) settings. We can see that X-AVDD significantly outperforms AVDD across most SPC settings. In the ConvNet architecture, at extremely low SPC, the extra cross-modal constraints can over-regularize, whereas benefits emerge as soon as synthetic diversity increases (SPC $\geq$ 10).
There is also a particularly significant jump across all SPCs for VGGSound, which is by far the largest dataset used in our experiments.
\begin{table}[ht!]
\centering
\caption{\small Comparison of accuracy (\%) on synthetic datasets for various architectures with the ensemble fusion method. For the ConvNet architecture, comparisons are carried out against DM and AVDD, while for ResNet-18 and ViT, comparisons are carried out against a multimodal implementation of SRe$^2$L. SPC denotes samples per class. Bold indicates best performance within each architecture group. - indicates Out-of-Memory (OOM) errors.}
\label{tab:combined_results}
\resizebox{\columnwidth}{!}{%
\begin{tabular}{ll c ccc c cc c cc}
\toprule
& & & \multicolumn{3}{c}{\textbf{ConvNet}} & & \multicolumn{2}{c}{\textbf{ResNet-18}} & & \multicolumn{2}{c}{\textbf{ViT}}\\ 
\cmidrule(r){4-6} \cmidrule(l){8-9} \cmidrule(l){11-12}
Dataset & SPC & Ratio (\%) & DM & AVDD & X-AVDD & & SRe$^2$L & X-AVDD & & SRe$^2$L & X-AVDD \\
\midrule
\textbf{VGGS-10K} & 1 & 0.11 & 36.5 $\pm$ 2.5 & \bf 40.4 $\pm$ 1.8 & 37.6 $\pm$ 0.8 & & 21.7 $\pm$ 0.6 & \bf 30.0 $\pm$ 1.5 & & 16.7 $\pm$ 0.9 & \bf 20.2 $\pm$ 0.5\\
 & 10 & 1.13 & 43.8 $\pm$ 1.7 & 55.0 $\pm$ 1.7 & \bf 57.2 $\pm$ 0.9 & & 46.9 $\pm$ 2.0 & \bf 54.5 $\pm$ 0.7 & & 25.4 $\pm$ 0.5 & \bf 30.0 $\pm$ 0.7\\
& 20 & 2.26 & 49.0 $\pm$ 2.4 & 58.0 $\pm$ 1.7 & \bf 62.7 $\pm$ 0.5 & & 52.7 $\pm$ 0.5 & \bf 61.5 $\pm$ 0.8 & & 26.1 $\pm$ 1.8 & \bf 35.9 $\pm$ 0.8\\
& 50 & 5.65 & - & - & \bf 66.1 $\pm$ 0.3 & & 60.4 $\pm$ 0.9 & \bf 66.7 $\pm$ 0.9 & & 30.2 $\pm$ 1.2 & \bf 42.1 $\pm$ 1.2\\
& Full & & \multicolumn{3}{c}{74.7} & & \multicolumn{2}{c}{77.4} & & \multicolumn{2}{c}{62.1} \\
\midrule 
\textbf{AVE} & 1 & 0.1 & 21.7 $\pm$ 1.5 & \bf 23.0 $\pm$ 1.4 & 22.0 $\pm$ 0.4 & & 18.5 $\pm$ 1.0 & \bf 20.3 $\pm$ 0.8 & & 11.2 $\pm$ 0.8 & \bf 12.4 $\pm$ 0.7\\
 & 10 & 1.0 & 28.1 $\pm$ 1.8 & 36.8 $\pm$ 0.9 & \bf 43.5 $\pm$ 0.2 & & 41.2 $\pm$ 0.5 & \bf 46.9 $\pm$ 0.7 & & 24.5 $\pm$ 0.7 & \bf 26.7 $\pm$ 0.1\\
& 20 & 2.0 & 32.6 $\pm$ 1.0 & 40.1 $\pm$ 1.0 & \bf 50.2 $\pm$ 0.8 & & 45.9 $\pm$ 0.3 & \bf 50.3 $\pm$ 0.3 & & 25.3 $\pm$ 2.3 & \bf 31.1 $\pm$ 0.8\\
& 50 & 5.0 & - & - & \bf 56.2 $\pm$ 0.6 & & 49.1 $\pm$ 0.3 & \bf 56.3 $\pm$ 0.6 & & 27.8 $\pm$ 1.3 & \bf 33.0 $\pm$ 0.4\\
& Full & & \multicolumn{3}{c}{58.5} & & \multicolumn{2}{c}{59.2} & & \multicolumn{2}{c}{47.9} \\
\midrule 
\textbf{MUSIC-21} & 1 & 0.014 & 38.3 $\pm$ 1.3 & 44.0 $\pm$ 2.2 & \bf 47.9 $\pm$ 0.7 & & 21.6 $\pm$ 0.8 & \bf 30.8 $\pm$ 2.1 & & 13.6 $\pm$ 1.3 & \bf 21.0 $\pm$ 0.5\\
 & 10 & 0.14 & 54.8 $\pm$ 1.4 & 68.1 $\pm$ 1.0 & \bf 74.5 $\pm$ 0.5 & & 58.7 $\pm$ 1.4 & \bf 70.9 $\pm$ 0.3 & & 37.4 $\pm$ 0.3 & \bf 44.5 $\pm$ 0.1\\
& 20 & 0.28 & 61.1 $\pm$ 1.3 & 70.3 $\pm$ 0.7 & \bf 77.6 $\pm$ 0.9 & & 61.2 $\pm$ 2.9 & \bf 74.1 $\pm$ 0.4 & & 41.6 $\pm$ 0.9 & \bf 48.0 $\pm$ 1.4\\
& 50 & 0.70 & - & - & \bf 79.7 $\pm$ 0.3 & & 71.9 $\pm$ 0.6 & \bf 78.4 $\pm$ 0.3 & & 52.2 $\pm$ 1.2 & \bf 56.7 $\pm$ 0.8 \\
& Full & & \multicolumn{3}{c}{80.5} & & \multicolumn{2}{c}{81.4} & & \multicolumn{2}{c}{71.3} \\
\midrule 
\textbf{VGGSound} & 1 & 0.18 & 3.1 $\pm$ 0.2 & 5.0 $\pm$ 0.3 & \bf 18.3 $\pm$ 0.3 & & 7.9 $\pm$ 0.9 & \bf 10.9 $\pm$ 0.2 \\
& 10 & 1.87 & 6.4 $\pm$ 0.3 & 8.2 $\pm$ 0.2 & \bf 24.0 $\pm$ 0.1 & & 17.3 $\pm$ 1.4 & \bf 22.4 $\pm$ 0.5 \\
& 20 & 3.74 & 8.6 $\pm$ 0.1 & 9.8 $\pm$ 0.3 & \bf 26.5 $\pm$ 1.1 & & 21.1 $\pm$ 0.3 & \bf 25.1 $\pm$ 0.1 \\
& 50 & 9.35 & - & - & \bf 28.2 $\pm$ 0.7 & & 22.2 $\pm$ 0.5 & \bf 28.4 $\pm$ 0.9 \\
& Full & & \multicolumn{3}{c}{30.1} & & \multicolumn{2}{c}{32.1} \\
\bottomrule
\end{tabular}}
\end{table}

\subsection{Results for AudioCLIP architecture}
AudioCLIP~\citep{9747631} extends CLIP to the Audio modality along with image and text, incorporating ESResNeXt audio-model into the CLIP framework. Table~\ref{Tab:audioclip_results} provides a comparative analysis of accuracy values obtained on the test set of VGGSound-10K and AVE datasets with synthetic images generated by our proposed method. With a fraction of the time the model takes to train on the full dataset, distilled images can achieve comparable generalization accuracy, underscoring the role of distillation in training an audio-visual dataset with SOTA architectures.

\begin{table}[hbt!]
\small
    \centering
        \caption{\small Test accuracy obtained by synthetic data generated by X-AVDD on VGGSound-10K and AVE datasets with AudioCLIP architecture. Ratio is the percentage of the condensed image as compared to the full dataset. With a fraction of the time taken to train the model on the full dataset, distilled images can achieve comparable generalization accuracy, underscoring the role of distillation for training audio-visual datasets with SOTA architectures.}
        \resizebox{\columnwidth}{!}{%
        \begin{tabular}{l c ccc c ccc}
        \toprule 
        & & \multicolumn{3}{c}{VGGSound-10K} & & \multicolumn{3}{c}{AVE} \\
        \cmidrule(r){3-5} \cmidrule(l){7-9}
        SPC & & Ratio (\%) & Accuracy (\%) & Train time (Hrs.) & & Ratio (\%) & Accuracy (\%) & Train time (Hrs.) \\ 
        \midrule 
        1 & & 0.11 & 24.8 $\pm$ 0.6 & 0.28 & & 0.1 & 44.0 $\pm$ 1.2 & 0.43 \\
        10 & & 1.13 & 33.7 $\pm$ 0.3 & 0.56 & & 1.0 & 74.5 $\pm$ 0.9 & 1.31 \\
        20 & & 2.26 & 46.8 $\pm$ 0.9 & 1.16 & & 2.0 & 76.8 $\pm$ 1.1 & 2.66 \\
        50 & & 5.65 & 62.3 $\pm$ 0.8 & 2.74 & & 5.0 & 81.3 $\pm$ 0.9 & 6.43 \\
        Full & & 100 & 72.4 $\pm$ 0.7 & 25.30 & & 100 & 85.6 $\pm$ 1.1 & 72.12 \\
        \bottomrule
        \end{tabular}}
        \label{Tab:audioclip_results}
\end{table}

\subsection{Cross architecture analysis}
To study the generalization capability of the generated synthetic dataset, we carried out a cross-architecture study with two different architectures compared to the base model through which the synthetic dataset is generated. We utilized VGG-11~\citep{simonyan2015deepconvolutionalnetworkslargescale}, and transformer-based architecture, ViT~\citep{dosovitskiy2021an}. The results are listed in Table~\ref{Tab:cross_arch_results}. Our proposed method significantly outperforms the baseline across various datasets and SPC settings. Significant improvement over the baseline on the MUSIC-21 dataset with ViT as the target architecture provides further proof of the efficacy of our proposed method.

\begin{table}[hbt!]
    \centering
\caption{\small Comparative analysis of cross-architecture generalization performance of the distilled dataset generated using the ResNet-18 architecture for both audio and video modality. We experimented with two target architectures: VGG11 and ViT. For both, the synthetic dataset generated by X-AVDD achieves higher generalization accuracy than that generated by the SRe$^2$L adaptation. Results are provided for the Ensemble fusion method.}
\resizebox{0.9\columnwidth}{!}{%
    \begin{tabular}{ll c cc c cc c cc}
    \toprule 
    & & & \multicolumn{2}{c}{VGGSound-10K} & & \multicolumn{2}{c}{AVE} & & \multicolumn{2}{c}{Music-21} \\ 
    \cmidrule(r){4-5} \cmidrule(l){7-8} \cmidrule(l){10-11}
    Target & SPC & & SRe$^2$L & X-AVDD  & & SRe$^2$L & X-AVDD  & & SRe$^2$L & X-AVDD \\
    \midrule 
    VGG-11 & 1 & & 14.6 $\pm$ 3.7 & \bf 20.4 $\pm$ 2.3 & & 6.2 $\pm$ 2.5 &  \bf 11.4 $\pm$ 0.7 & & 18.2 $\pm$ 1.9 & \bf 22.3 $\pm$ 1.2 \\ 
    & 10 & & 34.8 $\pm$ 1.3 & \bf 41.3 $\pm$ 2.2 & & 21.8 $\pm$ 0.4 & \bf 29.9 $\pm$ 0.7 & & 39.7 $\pm$ 1.6 & \bf 47.3 $\pm$ 2.6\\
    & 20 & & 44.0 $\pm$ 2.4 & \bf 51.5 $\pm$ 0.8 & & 29.1 $\pm$ 0.3 & \bf 35.6 $\pm$ 0.6 & & 42.8 $\pm$ 1.1 & \bf 58.7 $\pm$ 1.9\\ 
    & 50 & & 53.2 $\pm$ 0.2 & \bf 59.4 $\pm$ 0.2 & & 35.0 $\pm$ 0.7 & \bf 42.4 $\pm$ 0.3 & & 48.5 $\pm$ 0.5 & \bf 64.3 $\pm$ 1.1\\
    \midrule 
    ViT & 1 & & 13.9 $\pm$ 1.1 & \bf 23.3 $\pm$ 0.9 & & 12.6 $\pm$ 0.8 & \bf 13.9 $\pm$ 0.1 & & 10.0 $\pm$ 0.5 & \bf 14.4 $\pm$ 1.8\\
    & 10 & & 30.0 $\pm$ 2.9 & \bf 39.9 $\pm$ 0.8 & & 21.0 $\pm$ 0.4 & \bf 27.2 $\pm$ 1.1 & & 26.7 $\pm$ 2.0 & \bf 44.5 $\pm$ 0.8\\ 
    & 20 & & 38.6 $\pm$ 0.6 & \bf 46.2 $\pm$ 0.7 & & 23.1 $\pm$ 0.3 & \bf 30.4 $\pm$ 0.7 & & 28.8 $\pm$ 0.7 & \bf 45.1 $\pm$ 0.9\\
    & 50 & & 44.3 $\pm$ 1.1 & \bf 51.3 $\pm$ 1.4 & & 25.5 $\pm$ 2.4 & \bf 35.3 $\pm$ 0.5 & & 29.0 $\pm$ 0.9 & \bf 56.5 $\pm$ 1.3 \\ 
    \bottomrule
    \end{tabular}}
    \label{Tab:cross_arch_results}
\end{table}

\subsection{Transfer learning with frozen backbone of SOTA image and audio encoder models}
In this experiment, we investigate whether distilled datasets generated using simpler models such as ConvNet and ResNet-$18$, are sufficient to train lightweight classifiers on top of frozen, state-of-the-art image and audio encoder models. We used two complementary setups for this experiment. The first setup uses the audio and visual backbones of ImageBind~\citep{girdhar2023imagebind}, which are designed to map multimodal data onto a shared feature space. The second setup leverages Audio Spectrogram Transformer (AST)~\citep{gong2021ast} as the audio encoder and SigLIP2~\citep{tschannen2025siglip2} as the visual encoder. We chose AST because of its strong performance on music/environmental audio domains on MAEB~\citep{assadi2026maeb}, and chose SigLIP2~\citep{tschannen2025siglip2} because the SigLIP family is among the strongest open image embedding models on MIEB~\citep{xiao2025mieb}. MAEB and MIEB are integrated with the very popular MTEB benchmark~\citep{muennighoff2023mteb}. Small projection heads and linear classifiers are then trained on the extracted, frozen features using only the distilled synthetic data. 

We compare the performance of data distilled by AVDD and X-AVDD using ConvNet, and the data distilled by SRe$^2$L and X-AVDD using ResNet-$18$. Table~\ref{Tab:transfer_learning} tabulates these results for the VGGSound-10K dataset. With only $50$ samples per class, the dataset distilled by X-AVDD using ResNet-18 achieves significant performance compared to training with the full dataset, highlighting the capability of X-AVDD and its adaptation to powerful pretrained foundation models for downstream audio-visual classification tasks. 

\begin{table}[hbt!]
    \centering
\caption{\small Performance analysis of distilled dataset for transfer learning with frozen backbone of SOTA models. X-AVDD (C) and X-AVDD (R) denote the synthetic datasets generated by ConvNet and ResNet-$18$, respectively. With only $50$ samples per class, synthetic data generated by the ResNet-18 architecture and X-AVDD achieves performance comparable to that of training on the full dataset.}

\resizebox{\columnwidth}{!}{%
    \begin{tabular}{l c cccc c cccc}
    \toprule 
    & & \multicolumn{4}{c}{ImageBind} & & \multicolumn{4}{c}{AST + SigLIP2} \\ 
    \cmidrule(r){3-6} \cmidrule(l){8-11}
    SPC & & AVDD & X-AVDD (C) & SRe$^2$L & X-AVDD (R) & & AVDD & X-AVDD (C) & SRe$^2$L & X-AVDD (R) \\ 
    \midrule 
    1 & & 20.1 $\pm$ 0.4 & 25.7 $\pm$ 0.8 & 25.2 $\pm$ 0.6 & \bf 28.3 $\pm$ 0.4 & & 23.1 $\pm$ 1.3 & 24.3 $\pm$ 0.3 & 18.1 $\pm$ 1.2 & \bf 37.3 $\pm$ 0.5\\
    10 & & 26.4 $\pm$ 0.9 & 40.9 $\pm$ 0.6 & 40.1 $\pm$ 0.8 & \bf 58.4 $\pm$ 0.5 & & 36.1 $\pm$ 1.0 & 42.2 $\pm$ 0.7 & 49.8 $\pm$ 0.7 & \bf 58.7 $\pm$ 0.9\\
    20 & & 33.2 $\pm$ 0.5 & 43.1 $\pm$ 0.4 & 43.7 $\pm$ 0.3 & \bf 64.3 $\pm$ 0.5 & & 37.7 $\pm$ 1.3 & 49.1 $\pm$ 1.6 & 51.2 $\pm$ 1.2 & \bf 65.6 $\pm$ 0.6 \\
    50 & & - & 48.3 $\pm$ 0.3 & 56.7 $\pm$ 0.6 & \bf 67.1 $\pm$ 0.3 & & - & 51.7 $\pm$ 1.1 & 56.5 $\pm$ 0.7 & \bf 73.5 $\pm$ 0.3 \\
    Full & & \multicolumn{4}{c}{79.8} && \multicolumn{4}{c}{81.3} \\
    \bottomrule
    \end{tabular}}
    \label{Tab:transfer_learning}
\end{table}
\subsection{Memory comparison}
For a fair comparison, we ran both AVDD and X-AVDD on a single 32 GB Tesla V100 GPU for SPC=$20$ on ConvNet architecture.
AVDD must contain all synthetic images during the optimization stage, whereas X-AVDD can generate images batch-wise, drastically reducing memory requirements. Table~\ref{Tab:memory_compare} compares GPU-Hour consumption for generating synthetic datasets by AVDD and X-AVDD. 
This significant reduction in memory consumption enables X-AVDD to scale to higher SPC settings and larger architectures. 

\begin{table}[hbt!]
\centering
\caption{\small GPU-Hour comparison between AVDD and our proposed method for various datasets.}
\resizebox{0.5\columnwidth}{!}{%
\begin{tabular}{ccccc}
\toprule
Method & VGGSound-10K & AVE & MUSIC-21 & VGGSound \\
\midrule 
AVDD & 165 & 471 &  363 & 5342\\
X-AVDD & 23 & 51 & 45 & 628\\
\bottomrule
\end{tabular}}
\label{Tab:memory_compare}
\end{table}
\subsection{Ablation Studies}

\subsubsection{Impact of \texorpdfstring{$\alpha$}{alpha} on test accuracy}
\label{sec:alpha_test}
\begin{figure}[hbt!]
    \centering
    \includegraphics[width=0.48\columnwidth,trim={0 0 0 0}, clip]{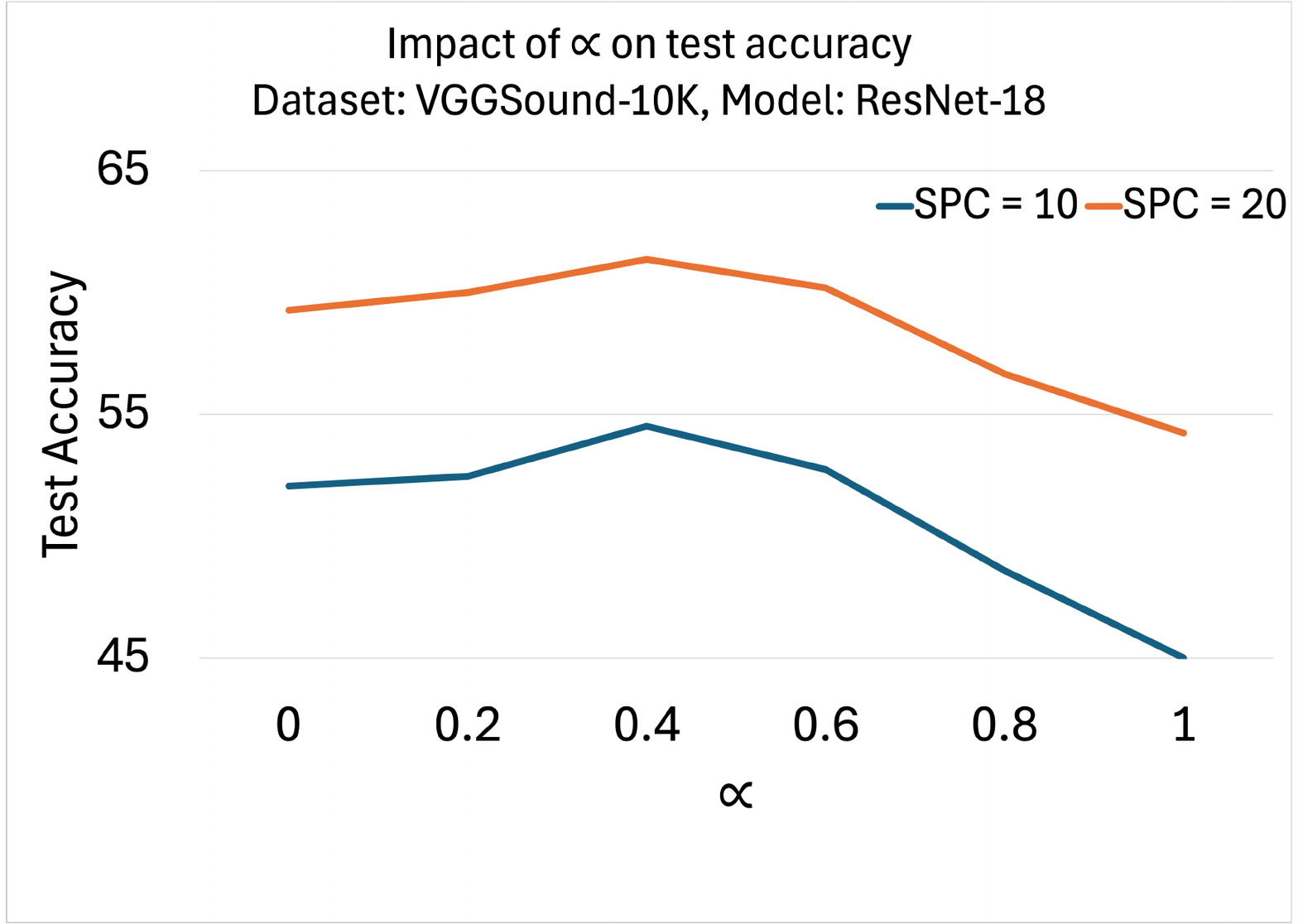}
    \hfill
    \includegraphics[width=0.48\columnwidth,trim={0 0 0 0}, clip]{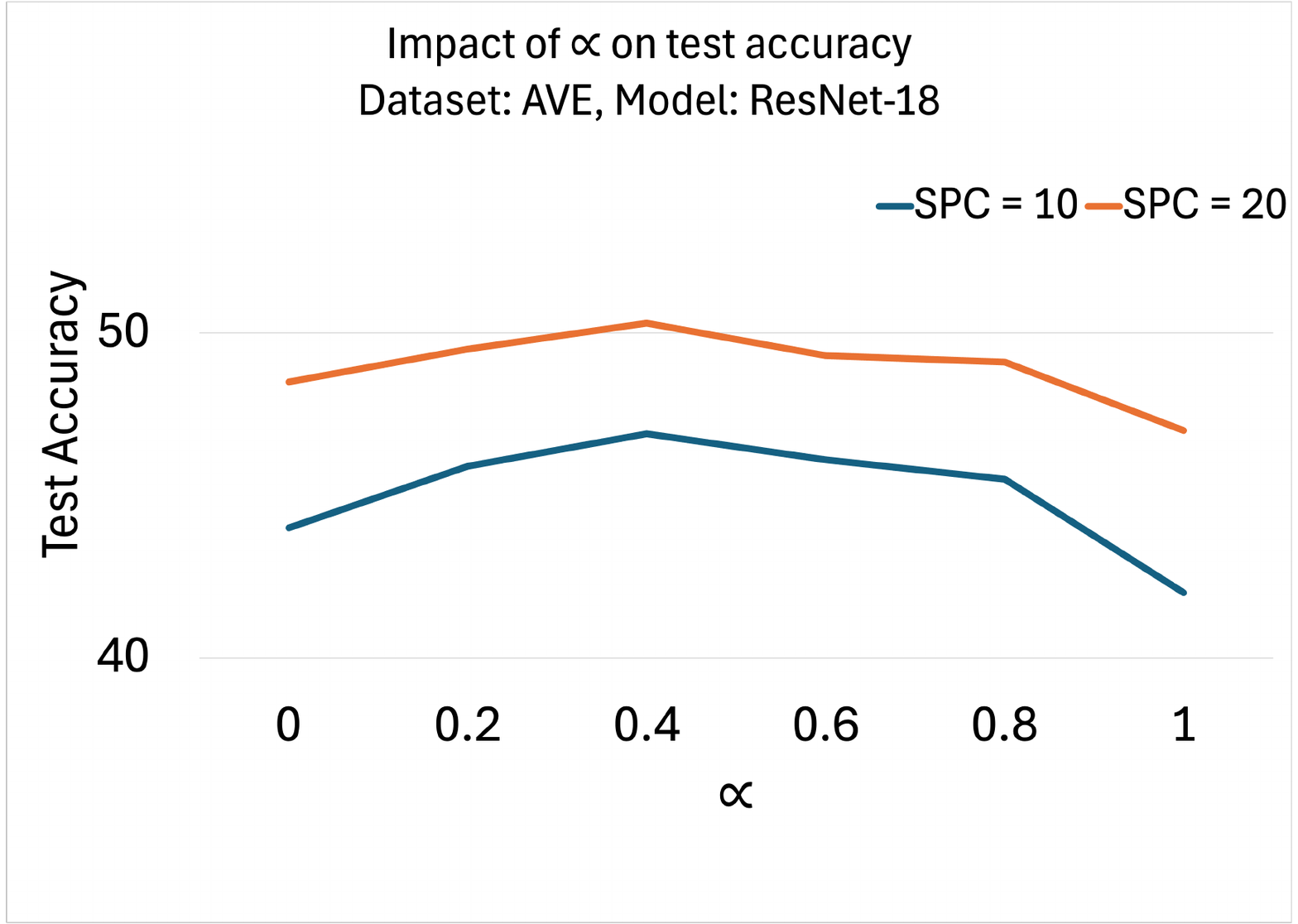}
    \caption{Impact of $\alpha$ on test accuracy. The left panel shows results for the VGGSound-$10$K dataset and the ResNet-18 architecture, while the right panel shows results for AVE and ResNet-18.} 
    \label{fig:alpha_ablation}
\end{figure}

$\alpha$ is the weight assigned to the cross-entropy loss from hard labels during the training of the student network with the synthetic dataset, where the total loss is a convex combination of cross-entropy loss from hard labels and KL divergence loss from soft labels. Figure~\ref{fig:alpha_ablation} shows that the combination of both hard labels and soft labels results in higher generalization accuracy while training a student model from scratch using the distilled synthetic dataset. 

\subsubsection{Directionality ablation}
We ablate the two cross-modal EDM terms by toggling the weights in Eq.~\ref{eq:total_loss_syn}. Using either the video$\rightarrow$audio term ($\lambda_{av}=0$) or audio$\rightarrow$video term ($\lambda_{va}=0$) in a standalone fashion consistently underperforms the full bi-directional objective across SPC regimes. This suggests that bidirectional cross-modal constraints are complementary and important for synthesizing informative paired audio-visual data. Table~\ref{tab:directionality} enumerates these observations on VGGSound-10K dataset.
\subsubsection{Improvement Over Random}
DD-Ranking~\citep{li2025ddrankingrethinkingevaluationdataset} introduced a metric called Improvement Over Random (IOR) to carry out a fair evaluation of the impact of a dataset distillation method. IOR is computed by: $\text{IOR} = \mathrm{Acc}_{\text{syn-any}} - \mathrm{Acc}_{\text{rdm-any}}$
, where $\mathrm{Acc}_{\text{syn-any}}$ and $\mathrm{Acc}_{\text{rdm-any}}$ are test accuracies of a model trained on the synthetic dataset and a randomly selected subset from the original dataset, respectively. 
Table~\ref{Tab:IOR_comparison} enumerates the IOR value for each dataset across each SPC setting. Our proposed method achieves significant improvement over random selection across multiple datasets.

\begin{table*}[t]
\centering
\begin{minipage}{0.56\textwidth}
\centering
\caption{\small Directionality ablation on VGGSound-10K. We toggle the EDM loss weights in Eq.~\ref{eq:total_loss_syn}.}
\resizebox{\linewidth}{!}{%
\begin{tabular}{lcccc}
\toprule
\textbf{Method} & \textbf{SPC=1} & \textbf{SPC=10} & \textbf{SPC=20} & \textbf{SPC=50} \\
\midrule
X-AVDD (both) & 30.0 {\textpm} 1.5 & 54.5 {\textpm} 0.7 & 61.5 {\textpm} 0.8 & 66.7 {\textpm} 0.9 \\
Only V$\rightarrow$A & 25.4 {\textpm} 1.3 & 50.7 {\textpm} 0.1 & 57.2 {\textpm} 0.6 & 63.1 {\textpm} 0.1 \\
Only A$\rightarrow$V & 25.9 {\textpm} 1.1 & 53.4 {\textpm} 0.6 & 58.5 {\textpm} 1.2 & 62.0 {\textpm} 0.9 \\
\bottomrule
\end{tabular}}
\label{tab:directionality}
\end{minipage}
\hfill
\begin{minipage}{0.42\textwidth}
\centering
\caption{\small IOR values for various SPC settings for different datasets.}
\resizebox{\linewidth}{!}{%
\begin{tabular}{lccc}
\toprule
Dataset & SPC=$1$ & SPC=$10$ & SPC=$20$ \\
\midrule
VGGSound-10K & 9.4 & 6.2 & 7.2 \\
Music-21 & 22.3 & 11.9 & 9.0 \\
AVE & 11.9 & 14.5 & 10.1 \\
\bottomrule
\end{tabular}}
\label{Tab:IOR_comparison}
\end{minipage}
\end{table*}

\subsubsection{Impact of soft label on generalization accuracy}
Table~\ref{Tab:hl_vs_sl} \&~\ref{Tab:hl_vs_sl_music_21} provides a comparative analysis between the accuracy obtained on training a student network with only hard labels vs. with a combination of hard and soft labels. We can see that training the student network with a combination of hard and soft labels improves test accuracy for both the SRe$^2$L and our proposed method. 

\begin{table}[hbt!]
\centering

\begin{minipage}{0.48\linewidth}
\centering
\caption{\small Impact of soft label on generalization accuracy (VGGSound-10K, ResNet-18).}
\resizebox{\columnwidth}{!}{%
\begin{tabular}{lccccc}
\toprule
SPC & Ratio (\%) & SRe$^2$L (HL) & SRe$^2$L (SL) & Ours (HL) & Ours (SL) \\
\midrule
1 & 0.11 & 16.5 {\textpm} 0.1 & 21.7 {\textpm} 0.6 & 24.5 {\textpm} 1.5 & 30.0 {\textpm} 1.5\\
10 & 1.13 & 31.4 {\textpm} 1.3 & 46.9 {\textpm} 2.0 & 44.9 {\textpm} 0.2 & 54.5 {\textpm} 0.7 \\
20 & 2.26 & 36.3 {\textpm} 2.1 & 47.6 {\textpm} 1.6 & 49.3 {\textpm} 0.9 & 61.5 {\textpm} 0.8\\
50 & 5.65 & 41.8 {\textpm} 0.7 & 60.4 {\textpm} 0.9 & 52.9 {\textpm} 0.3 & 66.7 {\textpm} 0.9\\ 
\bottomrule
\end{tabular}}
\label{Tab:hl_vs_sl}
\end{minipage}
\hfill
\begin{minipage}{0.48\linewidth}
\centering
\caption{\small Impact of soft label on generalization accuracy (MUSIC-21, ResNet-18).}
\resizebox{\columnwidth}{!}{%
\begin{tabular}{lccccc}
\toprule
SPC & Ratio (\%) & SRe$^2$L (HL) & SRe$^2$L (SL) & Ours (HL) & Ours (SL) \\
\midrule
1 & 0.014 & 14.1 {\textpm} 2.7 & 20.5 {\textpm} 1.0 & 21.6 {\textpm} 0.8 & 30.8 {\textpm} 2.1\\
10 & 0.14 & 37.1 {\textpm} 2.8 & 51.6 {\textpm} 1.1 & 58.7 {\textpm} 1.4 & 70.9 {\textpm} 0.3 \\
20 & 0.28 & 43.4 {\textpm} 0.5 & 56.1 {\textpm} 0.1 & 61.2 {\textpm} 2.9 & 74.1 {\textpm} 0.4\\
50 & 0.70 & 52.1 {\textpm} 0.6 & 61.6 {\textpm} 0.4 & 71.9 {\textpm} 0.6 & 78.4 {\textpm} 0.3\\ 
\bottomrule
\end{tabular}}
\label{Tab:hl_vs_sl_music_21}
\end{minipage}

\end{table}

\section{Conclusion}
In this work, we introduced X-AVDD, the first scalable audio-visual dataset distillation approach that explicitly uses bi-directional cross-attention between the two modalities to jointly synthesize synthetic audio-visual data. By distilling data through cross-attended embeddings rather than independently matched modality features, X-AVDD preserves semantic alignment between audio and visual streams while substantially reducing memory and computational overhead. 
Through extensive experiments, we demonstrated the superior performance of X-AVDD across four benchmark audio-visual datasets. We also illustrated the effectiveness of our method by cross-architecture comparison with VGG-11 and ViT. The distilled data generated by X-AVDD shows impressive performance in finetuning frozen backbones of SOTA models, further highlighting the efficacy of our proposed methodology. 

\textbf{Limitations \& Future work} The present paper focuses on audio-visual classification as the downstream task. Other downstream tasks, such as temporal localization and audio-visual retrieval, would be taken up as future work.

\clearpage
\bibliography{main}
\bibliographystyle{plainnat}

\clearpage
\appendix
\onecolumn

\section{Table of notation}
Table~\ref{table:notations} provides various notation used through the paper.
\begin{table}[hbt!]
\caption{Notation}
\centering
\begin{tabular}{ll}
    \toprule
    Symbol  & Description \\
    \midrule
    $\mathcal{V}$     & Original large-scale training dataset \\
    $\mathcal{S}$ & Distilled synthetic dataset \\
    $\tau_a$ & Tokenization operator for audio data \\
    $\tau_v$ & Tokenization operator for image data \\
    $\mathcal{T}_a$ & Audio $\rightarrow$ image branch of bi-directional cross-attention block \\
    $\mathcal{T}_v$ & Image $\rightarrow$ audio branch of bi-directional cross-attention block \\
    $\theta^\mathcal{S}$ & Combined network trained on synthetic dataset \\
    $\theta^\mathcal{V}$ & Combined network trained on original dataset \\

    $\eta$ & Learning rate for distillation \\
    $l_{ce}$ & Cross-entropy loss \\
    $\langle x,y \rangle$ & Inner product of vector $x$ and vector $y$ \\
    $\|x\|_2$ & $\ell_2$-norm (Euclidean-norm) of vector $x$ \\
    $\tau$ & Temperature parameter \\
    $\alpha$ & Weight assigned to hard label loss \\
    \bottomrule
\end{tabular}

\label{table:notations}
\end{table} 
\section{Diversity Comparison}
Following~\cite{dire2026wacv}, we use three different metrics for a comprehensive comparative analysis of the diversity of the generated synthetic images:
\begin{enumerate}
    \item \textbf{Vendi Score~\cite{friedman2023the}}, which is defined as the exponential of the Shannon entropy of the eigenvalues of a similarity matrix. A higher Vendi Score indicates greater diversity in the dataset. 
    \item \textbf{Cosine Similarity.} We compute intra-class pairwise cosine similarity among the embeddings of the synthetic dataset and use the mean as a measure of diversity. For a pair of embeddings $\{x,y\}$, their cosine similarity is defined as:
\begin{equation}
\displaystyle s_{cos}(x,y) = \frac{ \langle x,y \rangle }{ \|x\| \cdot \|y\|}
\label{eq:cosine_sim}
\end{equation}
A lower intra-class cosine similarity indicates higher diversity.

\item \textbf{Coverage~\cite{naeem2020reliable}} measures the fraction of real samples whose neighbourhoods contain at least one synthetic sample: 
\begin{equation}
\text{coverage} = \frac{1}{N} \sum_{i=1}^{N} 1_{\left\{ \exists\, j \ \text{s.t.} \ Y_j \in B\big(X_i, \text{NND}_k(X_i)\big) \right\}}
\label{eq:coverage}
\end{equation}
where, $X$ is the real dataset, $Y$ is the synthetic dataset, NND$_k$ represents $k^{th}$ Nearest Neighbor Distance, and $B(X,r)$ represents a ball of radius $r$ around the data point $X$. The distance computations here are carried out in the feature space. A higher coverage indicates greater diversity among the synthetic images. 
\end{enumerate}

\begin{table}[hbt!]
\centering
\caption{Comparative analysis of diversity metrics. Synthetic data generated by X-AVDD performs better across all three diversity metrics as compared to the baseline multimodal adaptation of SRe$^2$L.}

\begin{tabular}{lcccc}
\toprule
Dataset & Method & Vendi $\uparrow$ & Similarity $\downarrow$ & Coverage $\uparrow$\\
\midrule
VGGSound-10k & SRe$^2$L & 1.26 & 0.89 & 0.05 \\
& X-AVDD & \bf 1.65 & \bf 0.87 & \bf 0.15 \\
\midrule 
Music-21 & SRe$^2$L & 2.13 & 0.81 & 0.03 \\
& X-AVDD & \bf 3.04 & \bf 0.77 & \bf 0.13 \\
\midrule 
VGGSound & SRe$^2$L & 1.23 & 0.81 & 0.05 \\
& X-AVDD & \bf 1.71 & \bf 0.77 & \bf 0.11 \\
\bottomrule
\end{tabular}
\label{Tab:diversity_metric_comparison}
\end{table}

\section{Computing Infrastructure Information}
\label{sec:compute_hardware}
Information regarding computing infrastructure for carrying out the experiments is provided in Table~\ref{Tab:compute_information}.
\begin{table}[hbt!]
\centering
\caption{Details of the computing infrastructure.}
\begin{tabular}{lc}
\toprule
Description & Value \\
\midrule
GPU & 32 GB Tesla V100 \\
OS & Ubuntu 18.04.1 LTS \\
Python & 3.11.7 \\
Pytorch & 2.4.1 \\
NumPy & 1.24.3 \\
\bottomrule
\end{tabular}

\label{Tab:compute_information}
\end{table}

\section{Implementation of multimodal \texorpdfstring{SRe$^2$L}{SRe2L}}
\label{sre2l_implementation}
~\citet{yinsqueeze} introduced SRe$^2$L, a lightweight tripartite learning paradigm that can scale to larger datasets such as ImageNet-1K. The three stages of SRe$^2$L are:
\begin{itemize}
\item \textbf{S}queeze, in which a model is trained on the original training dataset, and the Batch-Norm (BN) layer statistics are stored.\item \textbf{Re}cover, in which the BN statistics are used to synthesize the condensed dataset.
\item \textbf{Re}label, in which the synthetic dataset is assigned soft labels through the model trained on the original dataset.
\end{itemize}

In this paper, we have adapted this methodology as a baseline for distilling audio-visual datasets. In the \textbf{S}queeze stage, a teacher network is trained on the full training dataset. The BN statistics stored in various BatchNorm layers in these networks are used during the synthesis stage. 

The synthesis stage (Recover) of SRe$^2$L consists of synthesizing the target data starting from independent random Gaussian noise. However, the random nature of Gaussian noise and its lack of semantic information are bottlenecks for effective information distillation. Hence, in our approach, we use real samples for initialization, as do some other dataset distillation methods~\citep{cazenavette2022distillation,du2024diversity}. For a fair comparison, the same initialization samples are used for this baseline and our proposed X-AVDD method.

\section{Dataset details}\label{appendix:dataset_details}
We use the same benchmark datasets and the corresponding preprocessing/splits from AVDD~\cite{kushwaha2024audiovisual}.

\begin{itemize}
    \item \textbf{AVE}~\citep{Tian_2018_ECCV}: The Audio-Visual Event (AVE) dataset contains 4,143 unconstrained 10-second YouTube videos spanning 28 event categories with temporally annotated audio-visual event boundaries at second-level granularity (each video contains at least one 2-second audio-visual event). Following AVDD, we segment each clip into non-overlapping 1-second audio-visual windows aligned with the annotations, resulting in train/val/test splits of 27,726 / 3,288 / 3,305 samples.

    \item \textbf{MUSIC-21}~\citep{Zhao_2019_ICCV}: MUSIC-21 is an expanded musical performance dataset containing 1,365 untrimmed videos of musical solos and duets; the original paper reports a split of 1,065 training videos and 300 test videos, with preprocessing into video shots (5,861 shots in total). Following AVDD, we use their processed split for the \emph{solo} subset, with train/val/test splits of 146,908 / 7,103 / 42,440 samples.

    \item \textbf{VGGSound}~\citep{9053174}: VGGSound is a large-scale audio-visual dataset of short YouTube clips curated to ensure audio-visual correspondence. We use the preprocessed train/test split provided by AVDD~\citep{kushwaha2024audiovisual}, consisting of approximately 165k / 13k train/test samples across 309 classes.

    \item \textbf{VGGSound-10K}: We additionally use the 10-class subset of VGGSound released with AVDD~\citep{kushwaha2024audiovisual}, which contains 8,808 / 444 train/test samples (using AVDD’s preprocessing and split).
\end{itemize}

\section{Hyperparameter details}\label{hyperparameter_details}
Table~\ref{Tab:squeeze_hyperparameter} provides the hyperparameters used during pre-training of the teacher network, audio-guided video network, and video-guided audio network. Table~\ref{Tab:synthesis_hyperparameter} tabulates the hyperparameters used during synthesis. 

\begin{table}[hbt!]
\centering

\begin{minipage}[t]{0.48\textwidth}
\setlength{\tabcolsep}{3pt}
\centering
\caption{Hyperparameter details for pre-training.}
\label{Tab:squeeze_hyperparameter}
\begin{tabular}{lc}
\toprule
Hyperparameter & Value \\
\midrule
Epochs & 100 \\
Optimizer & Adam \\
Base learning rate & 0.001 \\
Learning rate schedule & cosine decay \\
Augmentation & RandomResizedCrop \\
Batch size & 128 \\
\bottomrule
\end{tabular}
\end{minipage}
\hfill
\begin{minipage}[t]{0.48\textwidth}
\setlength{\tabcolsep}{3pt}
\centering
\caption{Hyperparameter details for synthesis.}
\label{Tab:synthesis_hyperparameter}
\begin{tabular}{lc}
\toprule
Hyperparameter & Value \\
\midrule
Iteration & 1000 \\
Optimizer & Adam \\
Optimizer momentum & $\beta_1,\beta_2=0.5,0.9$ \\
Base learning rate & 0.25 \\
Learning rate schedule & cosine decay \\
Augmentation & RandomResizedCrop \\
Attn.\ multiplier & 1.0 \\
\bottomrule
\end{tabular}
\end{minipage}

\end{table}

\section{Timing comparison}
For a fair comparison, we ran both AVDD and our proposed X-AVDD on a single 32 GB Tesla V100 GPU. It may be noted that the training of the teacher network is a one-time process. The synthesis stage of our X-AVDD takes a minuscule amount of time as compared to AVDD. 
\begin{table}[hbt!]
\centering
\caption{Timing comparison of AVDD and our X-AVDD on VGGSound-10k dataset for synthesis of SPC=$20$. Kindly note that, while the synthesis stage in AVDD is a recurring process, in our proposed method, training the teacher network and storage of the embeddings are one-time requirements, and only the synthesis stage is the recurring step for generating synthetic samples across various SPC settings.}

\begin{tabular}{lclc}
\toprule
\multicolumn{2}{c}{AVDD} & \multicolumn{2}{c}{X-AVDD} \\
\toprule

Steps & Time & Steps & Time \\
\midrule
Buffer creation& 1 hr & teacher training \& embedding storage & 65 mins. \\
synthesis & 9 hrs. & synthesis & 12 mins. \\
\bottomrule
Total &10 hrs. & Total & 1.3 hrs.\\
\bottomrule
\end{tabular}

\label{Tab:timing_compare}
\end{table}

\section{Modality gap on VGGSound-10K}
\begin{figure}[hbt!]
    \centering
    \includegraphics[width=0.6\columnwidth, trim={0 0 0 0}, clip]{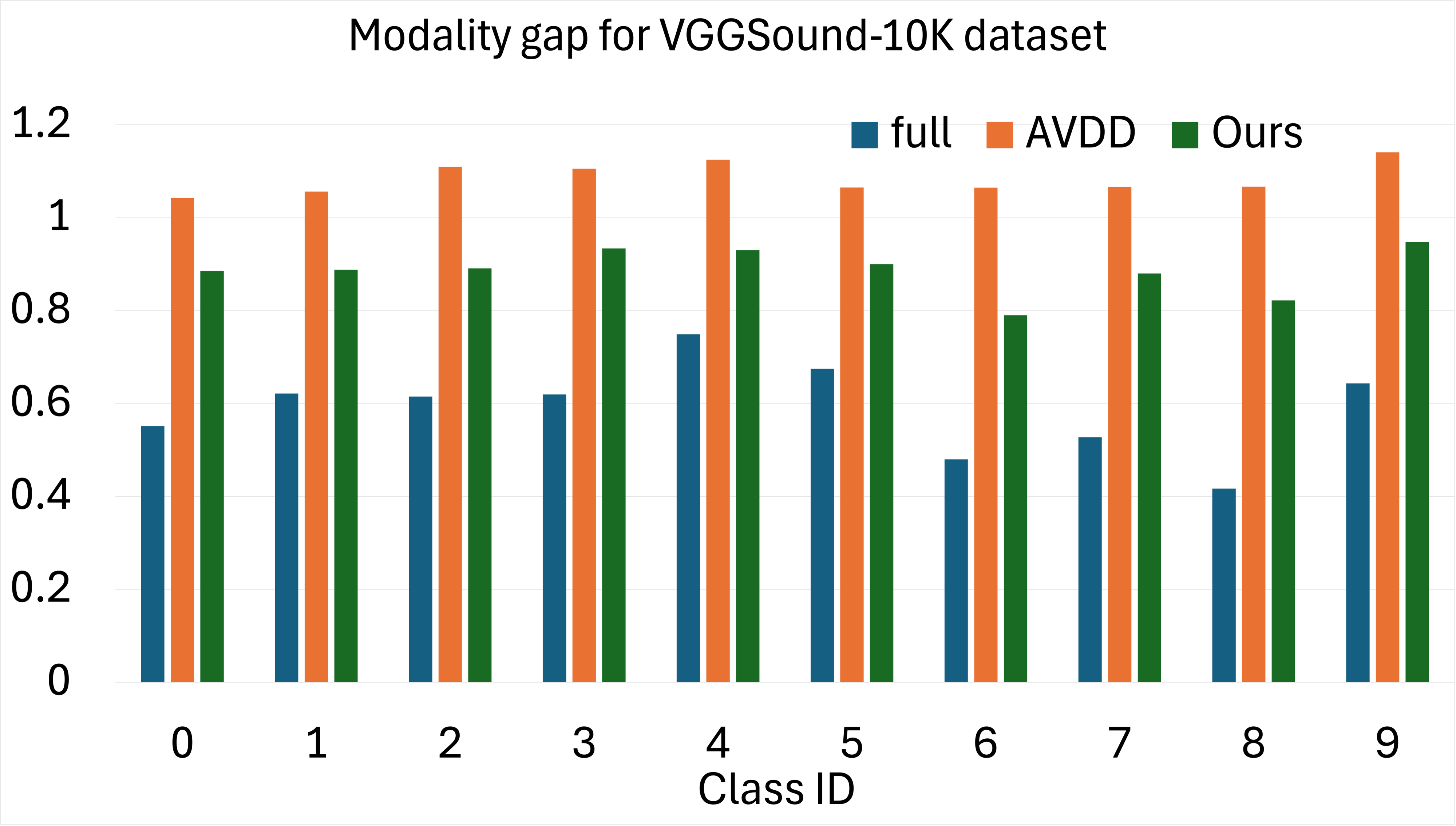}
    \caption{Comparison of modality gap observed in the real dataset, synthetic data generated by AVDD, and our proposed method. As shown, our proposed method achieves a lower modality gap consistently across all classes in the VGGSound-10K dataset.}
    \label{fig:modality_gap_vgg10k}
\end{figure}

Figure~\ref{fig:modality_gap_vgg10k} shows the significant increase in modality gap by synthetic data generated through AVDD on VGGSound-10K dataset. Our proposed method is able to achieve consistently lower modality gap across all classes.

\section{Impact of temperature on soft label generation}
\begin{figure}[hbt!]
    \centering
    \includegraphics[width=0.6\columnwidth, trim={0 0 0 0}, clip]{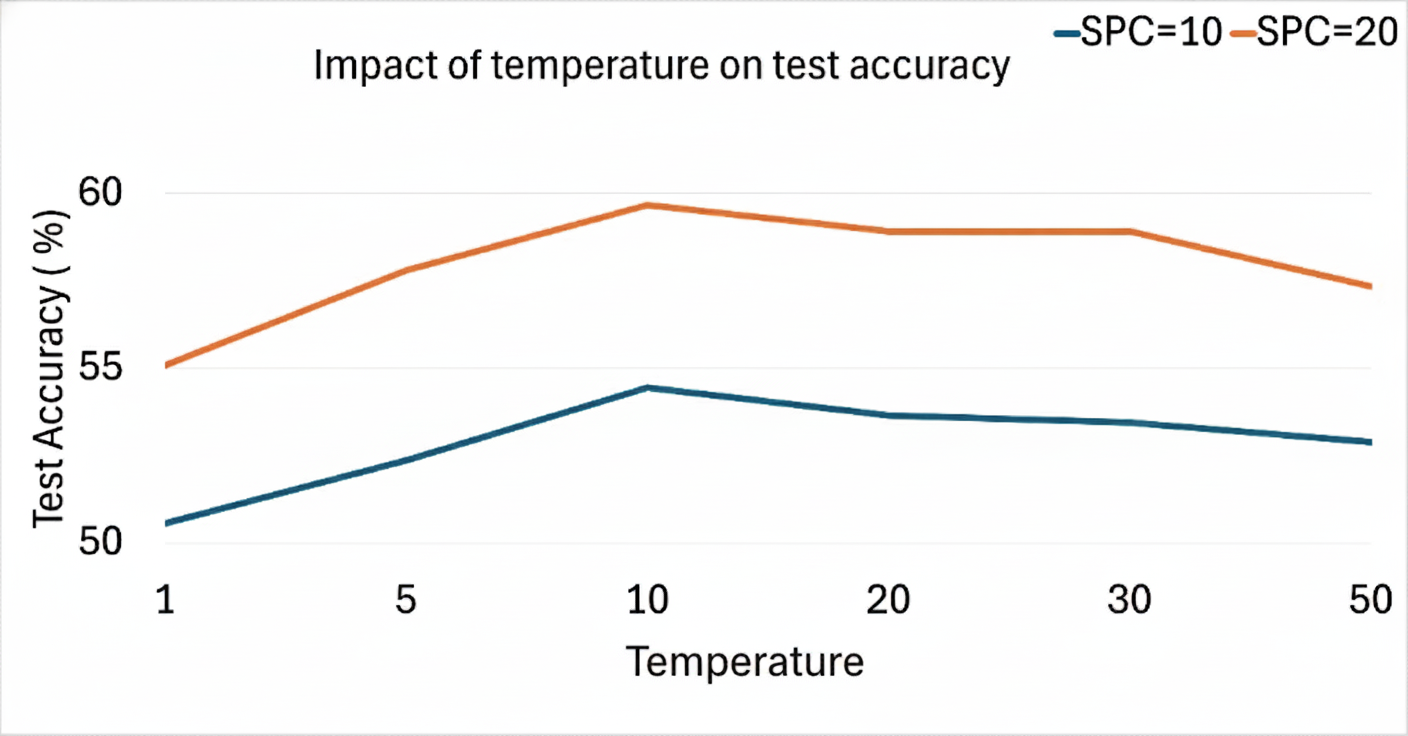}
    \caption{Impact of temperature on test accuracy. Results are provided for VGGSound-10K with ResNet-18 for SPC=$10$ and SPC=$20$. We can see that accuracy increases with temperature till it reaches a maximum and then decreases, signifying the importance of the temperature parameter during soft label generation.}
    \label{fig:temp_ablation}
\end{figure}

The temperature parameter $\tau$ is vital in smoothening the soft labels~\cite{label_smoothing_nips_2019}. The impact of $\tau$ on test accuracy for the VGGSound-10K dataset with the ResNet-18 architecture for SPC=$10$ and SPC=$20$ is depicted in Figure~\ref{fig:temp_ablation}.

\clearpage
\section{Results with `concat' fusion method}
The main paper contains results with the `ensemble' approach of audio-visual fusion. Table~\ref{Tab:concat_results} compares results obtained by the `concat' fusion method with a synthetic dataset generated by the SRe$^2$L baseline and X-AVDD methods. X-AVDD consistently outperforms the baseline across all four datasets and various SPC settings. 

\begin{table}[hbt!]
\centering
\caption{Test set accuracy comparison for various datasets with `concat' fusion method.}
\resizebox{0.5\columnwidth}{!}{%
\begin{tabular}{lccc}
\toprule
Dataset & SPC=$1$ & SRe$^2$L & X-AVDD \\
\midrule
VGGSound-10K & 1 & 20.8 $\pm$ 1.5 & \bf 26.8 $\pm$ 0.7 \\
& 10 & 47.2 $\pm$ 0.7 & \bf 55.4 $\pm$ 0.7 \\
& 20 & 49.6 $\pm$ 0.2 & \bf 65.1 $\pm$ 0.2 \\
& 50 & 64.4 $\pm$ 1.2 & \bf 69.5 $\pm$ 0.3 \\
\midrule 
AVE & 1 & 18.0 $\pm$ 0.1 & \bf 21.2 $\pm$ 0.4 \\ 
& 10 & 41.5 $\pm$ 0.6 & \bf 51.6 $\pm$ 0.4 \\
& 20 & 46.0 $\pm$ 0.7 & \bf 56.6 $\pm$ 0.2 \\
& 50 & 49.2 $\pm$ 0.4 & \bf 61.3 $\pm$ 0.2 \\
\midrule 
Music-21 & 1 & 29.8 $\pm$ 1.1 & \bf 34.0 $\pm$ 1.6 \\
& 10 & 66.7 $\pm$ 2.4 & \bf 70.8 $\pm$ 1.2 \\
& 20 & 67.1 $\pm$ 0.5 & \bf 73.5 $\pm$ 0.7 \\
& 50 & 72.1 $\pm$ 0.9 & \bf 79.0 $\pm$ 0.3 \\

\midrule 
VGGSound & 1 & 8.1 $\pm$ 1.2 & \bf 11.4 $\pm$ 0.9 \\
& 10 &  17.9 $\pm$ 0.8 & \bf 24.9 $\pm$ 0.3 \\
& 20 & 22.3 $\pm$ 0.1 & \bf 28.3 $\pm$ 0.2 \\
& 50 & 22.4 $\pm$ 0.6 & \bf 31.1 $\pm$ 0.1 \\ 

\bottomrule
\end{tabular}}

\label{Tab:concat_results}
\end{table}

\end{document}